\documentclass[10pt,twocolumn,letterpaper]{article}

\usepackage{cvpr}
\usepackage{algorithm}
\usepackage{algorithmic}
\usepackage{xcolor}
\usepackage{booktabs}
\usepackage{amsfonts}
\usepackage{hyperref}
\usepackage{comment}
\usepackage{textcomp}
\usepackage{listings}
\usepackage{adjustbox}
\usepackage{multirow}
\usepackage{multicol}
\usepackage{xcolor} 
\usepackage{subcaption}
\usepackage{amsmath}
\usepackage[toc,page]{appendix}
\usepackage{amssymb}
\usepackage{listings}

\title{CorrGAN: Input Transformation Technique Against Natural Corruptions}
\def\cvprPaperID{XXXX} 
\def\confName{CVPR}
\def\confYear{2022}

\author{Mirazul Haque\\
UT Dallas\\
{\tt\small mirazul.haque@utdallas.edu}
\and
Christof J. Budnik\\
Siemens Corporation, Technology\\
{\tt\small christof.budnik@siemens.com}
\and
Wei Yang\\
UT Dallas\\
{\tt\small wei.yang@utdallas.edu}

}
\begin{document}

\maketitle
\begin{abstract}

Because of the increasing accuracy of Deep Neural Networks (DNNs) on different tasks, a lot of real times systems are utilizing DNNs. These DNNs are vulnerable to adversarial perturbations and corruptions. Specifically, natural corruptions like fog, blur, contrast etc can affect the prediction of DNN in an autonomous vehicle. In real time, these corruptions are needed to be detected and also the corrupted inputs are needed to be de-noised to be predicted correctly. In this work, we propose CorrGAN approach, which can generate benign input when a corrupted input is provided. In this framework, we train Generative Adversarial Network (GAN) with novel intermediate output-based loss function. The GAN can denoise the corrupted input and generate benign input. Through experimentation, we show that up to 75.2\% of the corrupted misclassified inputs can be classified correctly by DNN using CorrGAN. 

\end{abstract}
\section{Introduction}

Deep Neural Networks (DNNs) are being used in real-time tasks such as object recognition, natural language processing, speech processing, etc. Many of these tasks are being used to create intelligent and autonomous systems. Specifically, tasks like object detection are being used in autonomous vehicles, mobile phones, and health diagnostics. One common requirement of those systems is their need for industrial-grade robustness to operate within safety boundaries regardless of environmental conditions.   

The robustness of a DNN can be defined as the change of DNN prediction if a minimal amount of perturbation is added to the input. Significant number of works have been performed to evaluate the robustness of DNNs through devising adversarial attacks~\cite{carlini2016hidden,szegedy2013intriguing,chen2022nicgslowdown,haque2020ilfo}. The generated adversarial images for object recognition are synthesized by adding perturbation to benign images. However, for real-time systems, devised adversarial examples can only be a threat if they are represented through a printed image. Because of this reason, the scope of adversarial attacks against real-time systems becomes limited.

However, natural corruptions are not needed to be represented through a printed image. The natural corruptions can be defined as the natural phenomenon that would degrade the quality of the image captured by the real-time systems for object recognition. Because of the noise added to the images due to these natural corruptions, the object recognition model can mispredict the output.
As these corruptions are added to the image captured by real-time system's camera, these corruptions can be a larger threat for real-time systems than synthesized adversarial inputs. On the manufacturing shop floor, corruptions would lower the classification accuracy by causing a robot malfunction that can lead to damage to equipments or even harm to humans.

During inference time, we can handle corrupted inputs in two ways: input detection and input conversion. 
A substantial amount of research has been performed to detect out-of-distribution examples (\textit{e.g.,} naturally corrupted examples). However, for a real-time system, detection is not enough because systems like autonomous vehicles need correct predictions to function properly. Because of that reason, input conversion is required for naturally corrupted examples to correctly predict the output. 

 Current input conversion techniques \cite{samangouei2018defense} focus on generating inputs whose euclidean distance (in terms of pixels) is nearest to the original benign inputs. However,
decreasing the euclidean distance between generated and original input does not make sure that the predicted output of the object recognition model would be correct. As we have seen earlier from the adversarial attacks \cite{carlini2016hidden,szegedy2013intriguing}, because of minimal input distance DNNs can mispredict the output. 

Therefore, we propose CorrGAN\footnote{The work has been performed during internship at Siemens.}, which includes the DNN intermediate output values into the loss function. CorrGAN has two components. The first component is Noise Classifier (NC), which differentiates between various types of corruption. The second component is Generative Adversarial Network (GAN), which converts the corrupted input into a benign one. As different types of corruptions need different GANs for benign input generation, the NC's objective is to find the correct GAN for denoising.
Through evaluation, we find out that the denoiser successfully can convert up to 75.2\% of the misclassified inputs to correctly classified inputs. Also, we find that NC can achieve 97.47\% accuracy in predicting noise type.

\section{Background and Related Works}

\subsection{Natural Corruptions}
Corruption techniques \cite{hendrycks2019robustness} contain different visual corruption, which includes practical corruptions like fog, blur, brightness etc. There are 19 different corruption types we have used. For each corruption type, five corruption levels are created from severity level one to five, resulting in a total of 95 different visual corruptions. The noise added to the inputs by corruption techniques is humanly perceptible.


\subsection{Generative Adversarial Networks} 
\label{subsec:GAN}
Generative Adversarial Networks (GANs) are generative models which have been used frequently to augment different types of data like image, text, code, speech etc. GANs mainly consist of two components. First component is called generator $\mathcal{G}(\cdot)$ and the other component is discriminator $\mathcal{D}(\cdot)$.
The input $x$ of the generator $\mathcal{G}(\cdot)$ is a seed sample (generally a noisy sample) and the output $\mathcal{G}(x)$ is an in-distribution input (Input which belongs to the distribution of training data).
After generating the output to the original seed, the test sample $\mathcal{G}(x)$ is sent to the discriminator.
The discriminator $\mathcal{D}(\cdot)$ is designed to distinguish the generated test samples $\mathcal{G}(x)$ and the original in-distribution training samples $x$. 
After training, the generator would generate more in-distribution inputs; correspondingly, the discriminator would also be more accurate in distinguishing original samples and generated samples. After being well trained, the discriminator and the generator would reach a Nash Equilibrium, which implies the generated test samples are challenging to be distinguished from the in-distribution samples. The loss function for a traditional GAN can be referred as,

$\mathcal{L}_{GAN} = \mathbb{E}_{x} log\mathcal{D}(x) + \mathbb{E}_{x} log[1 - \mathcal{D}( \mathcal{G}(x))]$

\subsection{Adversarial Sample Detection}
For detecting adversarial samples, Hendrycks \textit{et al.} \cite{hendrycks2016baseline} propose a baseline to detect out-of-distribution samples using softmax value. Lee \textit{et al.} \cite{lee2018simple} propose mahalanobis distance based confidence score using hidden feaures to detect out-of-distribution and adversarial samples. GraN \cite{lust2020gran} detects the out-of-distribution samples by using weight gradient values. Yap \textit{et al.} \cite{yap2019detecting} propose to use entropy of salient maps to detect adversarial samples. However, for real time systems, input detection is not enough, input conversion is also needed.

\section{Approach}
In this section, first we discuss about the problem formulation and then elaborate the approach.  
\subsection{Problem Formulation}

CorrGAN is designed based on the paradigm of Generative Adversarial Networks (GANs), which was discussed earlier (Section \ref{subsec:GAN}). As discussed earlier, GANs mainly consist of a generator $\mathcal{G}(\cdot)$ and a discriminator $\mathcal{D}(\cdot)$.
However, for the purpose of denoising, we consider that the input $x$ of the generator $\mathcal{G}(\cdot)$ is a noisy corrupted image and the output $\mathcal{G}(x)$ is a perturbation.
Our objective is to generate denoised input $x+\mathcal{G}(x)$ after applying the generated perturbation to the noisy input.

For the purpose of denoising, we need to create a sample that not only belongs to the same data distribution but also is similar to the original image. Also, we need to ensure that the generated denoised example would have similar predictions as original image. Decreasing only the pixel difference between original and generated image might not be sufficient for that \cite{samangouei2018defense,yan2017dcgans}.

For example, let's assume that we want to generate a sample $x+\mathcal{G}(x)$, where the euclidean pixel difference between $x+\mathcal{G}(x)$ and original image $x_{orig}$ is $\delta$, where $\delta \rightarrow 0$. However, for many scenarios, the value of $\delta$ won't be 0, and there will be a small pixel difference between two images. As we know from the adversarial attacks that a small pixel difference can be cause for wrong prediction \cite{carlini2016hidden}, the challenge about small input pixel difference is needed to be addressed. 

To address this challenge, we propose to include the certainty of the DNNs into the loss function~\cite{lee2018simple}. It has been observed that hidden states of a DNN can be analyzed to decide model certainty. Therefore, we propose to include the difference between hidden layer outputs of original image and generated image in the loss function. Also, to denoise the corruption, we propose to generate perturbation instead of generating benign input directly. The benign input can be generated by adding perturbation with corrupted input.
So the total loss function would be,

\vspace{5px}

$\mathcal{L}_{total} =  \mathcal{L}_{GAN}+
    ||(x + \mathcal{G}(x))-x_{orig}|| 
    + ||Hid(x)+
    \mathcal{G}(x))-Hid(x_{orig})||$
\vspace{5px}

Hid represents the hidden layer output values of DNN.

However, training a single $\mathcal{G}(\cdot)$ is not enough to denoise all types of noises because the effect of different noises on benign inputs is generally different. For example, the perturbation needed to denoise low contrast images would not be similar to the perturbation needed to denoise images affected with glass blur noise. Because of that reason, we need to train different $\mathcal{G}(\cdot)$ for different noises. However, given a noisy image, it would be challenging to know which $\mathcal{G}(\cdot)$ to use for denoising. Therefore, we need to create a noise classifier too.

\subsection{Overview}
There are three components of the approach: \textit{Training the Generator}, \textit{Training the Noise Classifier} and, \textit{Architecture Overview}. 
\subsubsection{Training the Generator}
As mentioned earlier, to train the generator, we use Euclidean distance between hidden states of corrupted and benign inputs, along with Euclidean distance between pixel values of corrupted and benign inputs. We consider specific hidden state layers in this scenario. For example, ResNet-18 models~\cite{he2016deep} has total 18 residual blocks in model architecture. These 18 blocks are divided into four groups. We consider outputs of each group to include in loss function. Therefore, we define a hidden state values $Hid$ as,

\vspace{5px}

$Hid(x)=Hid_{G_1}(x)+Hid_{G_2}(x)+Hid_{G_3}(x)+Hid_{G_4}(x)$

\vspace{5px}

where $x$ is the  input, $G_N$ is the group number input.

\subsubsection{Training the Noise Classifier}
Noise Classifier (NC) is a classification model, which can predict which type of noise is added to the input given the input. Each noise adds a specific type of pattern to the benign inputs. For example, the snow corruption would add specific white patches to the input. Hence, a convolution-based classifier model would be able to detect the specific noise if the model is trained. 


\subsubsection{Architecture Overview}

\begin{figure}[t]%
	\centering
	\includegraphics[width=0.46\textwidth]{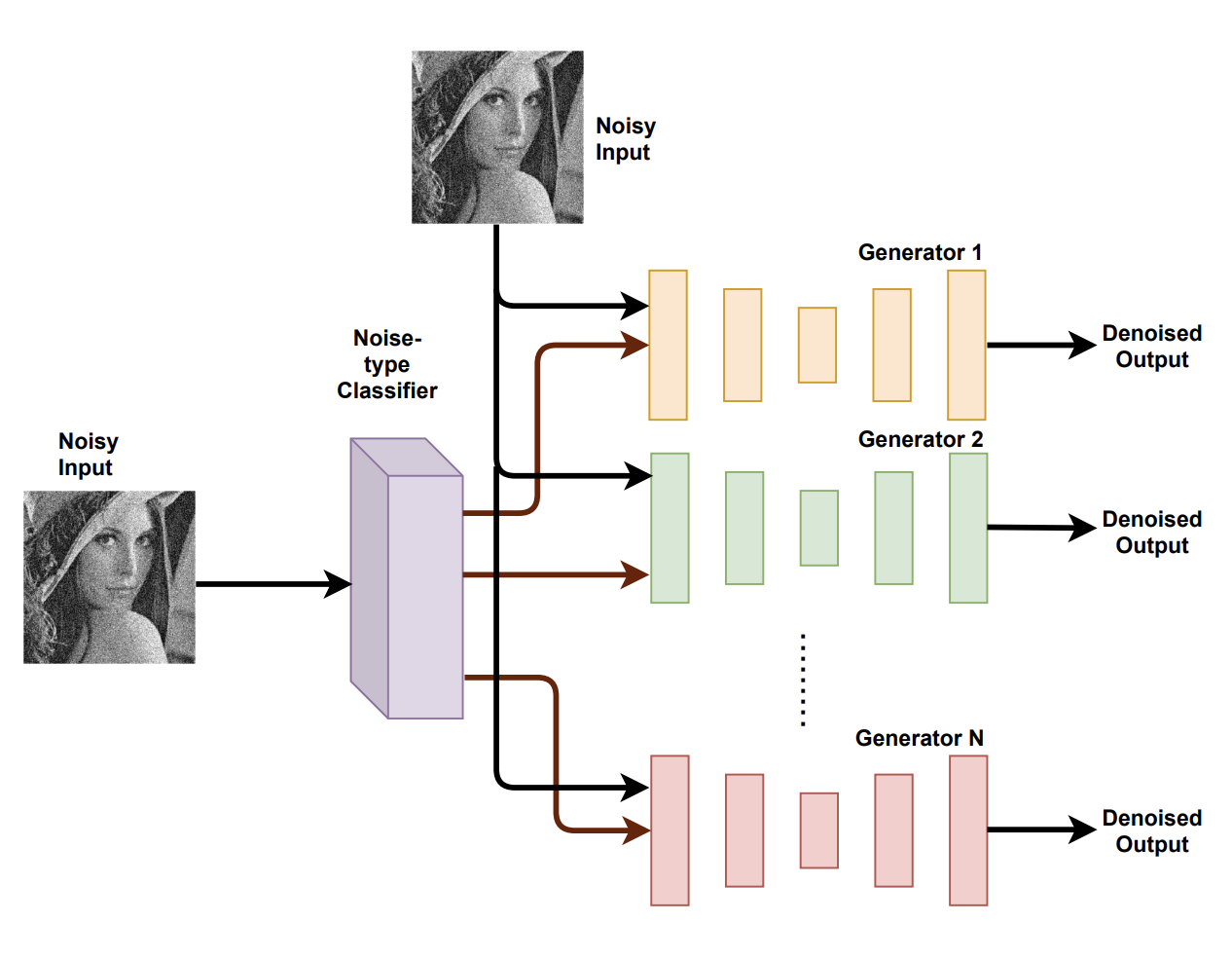}
    \caption{Working mechanism of CorrGAN}
	\label{fig:BD}
\end{figure}

Figure \ref{fig:BD} represents the overview of CorrGAN. If an input is rejected by the OOD detection system, it will be sent to Noise Classifier. The NC will predict which type of corruption is present in the input. Depending on the NC output, the denoiser GAN that is specified for the corruption is used to denoise the corrupted input.

\section{Evaluation}

We evaluate CorrGAN based on two perspective:

\textbf{1. What is the increase accuracy by CorrGAN against baseline?}

\textbf{2. What is the effectiveness of Noise Classifier?}

\subsection{Experimental Setup}
\textbf{Datasets.} We use CIFAR-10  dataset for the evaluation of CorrGAN. The dataset has 50,000 training images and 10,000 test images. Each type of corruption is added to the training and testing images with various scales.

\textbf{Models.} We have used ResNet-18 model~\cite{he2016deep} for the experimentation. 


\textbf{Baseline.} We use Defense-GAN \cite{samangouei2018defense} technique as baseline. 


\textbf{Threat Model.} In this work, we consider that we already know an input has been predicted incorrectly. As we mentioned in related works, multiple works has been performed about adversarial sample detection. Once the sample is detected, then the sample is fed to Noise Classifier (NC), and later to Generator model.

\begin{figure}[t]%
	\centering
	\includegraphics[width=0.35\textwidth]{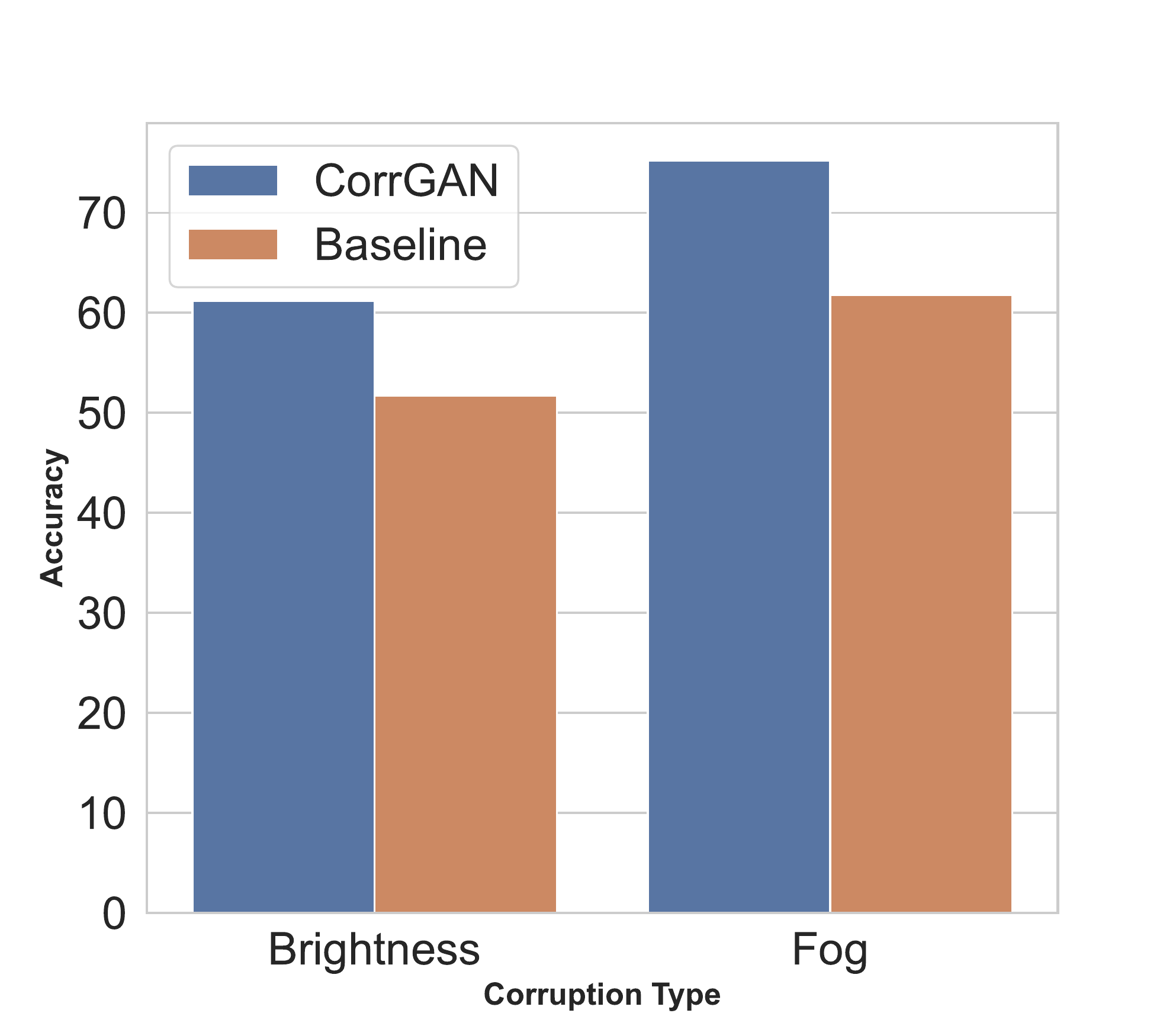}
    \caption{Comparison of Accuracy of CorrGAN generated inputs and Baseline generated input}
	\label{fig:acc}
\end{figure}

\subsection{Increasing Accuracy} We show the effectiveness of CorrGAN through two types of corruptions: Fog and Brightness. Both natural corruptions are some prevailing threats to industrial manufacturing processes.

First, we train one GAN for each type of corruption. The training data consists of 250,000 images. Similarly we train two separate defense-GANs for each type of corruption.

For testing, we first find out misclassified input by DNN (for each type of corruption) from 50,000 test inputs, and those selected inputs are fed to CorrGAN and baseline. Figure \ref{fig:acc} shows the result. It can be noticed that in both cases, the accuracy of CorrGAN generated inputs is more than 10\% higher than baseline generated inputs.


\subsection{Noise Classifier Accuracy} For testing the NC, we use a ResNet-18 model. The classes of this model represent different corruptions (fog and brightness in this scenario). We trained the ResNet-18 model given 250,000 corrupted inputs for each label and tested on 100,000 test samples. The model performs significantly well by achieving 97.47\% accuracy.  


\section{Application}
Object recognition is an important integral part of ensuring a safe operation for industrial manufacturing. In the future manufacturing shop floor, Self-Driving Vehicles (SDVs) and Autonomous Mobile Robots (AMRs) will be utilized to allow for more flexibility and increased productivity. Camera systems are supposed to monitor and detect the autonomous machines and humans operating in an unstructured and fenceless environment. Their detection will ensure to avoid any harm to their human co-workers by calculating their distances to each other.

The prerequisite for the safety of such an AI-controlled manufacturing system is the robustness to recognize the autonomous machines and workers on the shop floor. The detection by the camera system can, however, be hampered by natural phenomena that can corrupt the vision reception. On a factory floor, natural corruptions can include fog from steam sanitizing and disinfecting fogging or smoke as well as brightness from changing light conditions. Those natural corruptions add noise to the image, which needs to be identified and denoised.

Thus, the robustness of object detection of autonomous vehicles and humans on the manufacturing shop floor has become a major concern that needs to be addressed in order to become a widespread application in industry.
\section{Conclusion}

In this work, we propose CorrGAN, which uses intermediate output layers of DNN to generate benign outputs given corrupted inputs. We propose an architecture, which first identifies the type of corruption added to the input and then classifies it first. Next, a denoiser GAN would convert the corrupted input to benign input. Through evaluation, we find that CorrGAN can generate more accurate benign inputs than baseline technique.
\bibliographystyle{ieee_fullname}
\bibliography{egbib}
\end{document}